\documentclass[sigconf, nonacm]{acmart}
\AtBeginDocument{%
  }

\usepackage{algorithm}
\usepackage{algorithmic}
\usepackage{multirow}
\usepackage{enumitem}
\usepackage{pifont}
\usepackage{subcaption}
\usepackage{cuted} 
\usepackage{caption}

\setcopyright{acmlicensed}
\copyrightyear{2026}
\acmYear{2026}
\acmDOI{XXXXXXX.XXXXXXX}
\acmISBN{978-1-4503-XXXX-X/2026/08}

\begin{document}

\title{TraceMem: Weaving Narrative Memory Schemata from User Conversational Traces}


\author{Yiming Shu}
\affiliation{%
  \institution{The University of Hong Kong}
  \city{Hong Kong}
  \country{China}}
  \email{yiming.shu@connect.hku.hk}

\author{Pei Liu}
\affiliation{%
  \institution{The Hong Kong University of Science
and Technology (Guangzhou)}
  \city{Guangzhou}
  \country{China}}
  \email{pliu061@connect.hkust-gz.edu.cn}

\author{Tiange Zhang}
\affiliation{%
  \institution{Nankai University}
  \city{Tianjin}
  \country{China}}
  \email{2211123@mail.nankai.edu.cn}

\author{Ruiyang Gao}
\affiliation{%
  \institution{The University of Hong Kong}
  \city{Hong Kong}
  \country{China}}
  \email{gaoruiyang@connect.hku.hk}

\author{Jun Ma}
\affiliation{%
  \institution{The Hong Kong University of Science
and Technology (Guangzhou)}
  \city{Guangzhou}
  \country{China}}
  \email{jun.ma@ust.hk}

\author{Chen Sun}
\affiliation{%
  \institution{The University of Hong Kong}
  \city{Hong Kong}
  \country{China}}
  \email{c87sun@hku.hk}

\renewcommand{\shortauthors}{Trovato et al.}

\begin{abstract}
Sustaining long-term interactions remains a bottleneck for Large Language Models (LLMs), as their limited context windows struggle to manage dialogue histories that extend over time. Existing memory systems often treat interactions as disjointed snippets, failing to capture the underlying narrative coherence of the dialogue stream. We propose TraceMem, a cognitively-inspired framework that weaves structured, narrative memory schemata from user conversational traces through a three-stage pipeline: (1) Short-term Memory Processing, which employs a deductive topic segmentation approach to demarcate episode boundaries and extract semantic representation; (2) Synaptic Memory Consolidation, a process that summarizes episodes into episodic memories before distilling them alongside semantics into user-specific traces; and (3) Systems Memory Consolidation, which utilizes two-stage hierarchical clustering to organize these traces into coherent, time-evolving narrative threads under unifying themes. These threads are encapsulated into structured user memory cards, forming narrative memory schemata. For memory utilization, we provide an agentic search mechanism to enhance reasoning process. Evaluation on the LoCoMo benchmark shows that TraceMem achieves state-of-the-art performance with a brain-inspired architecture. Analysis shows that by constructing coherent narratives, it surpasses baselines in multi-hop and temporal reasoning, underscoring its essential role in deep narrative comprehension. Additionally, we provide an open discussion on memory systems, offering our perspectives and future outlook on the field. Our code implementation is available at: https://github.com/YimingShu-teay/TraceMem




\end{abstract}




\maketitle

\section{Introduction}
Despite broad competencies, Large Language Models (LLMs) struggle in long multi-turn dialogues due to their limited context window~\cite{zhong2024memorybank, sun2025scaling, zhang2025learn}. This bottleneck prevents them from building upon historical context, undermining dialogue coherence~\cite{liu2026simplemem, pan2025secom}. Therefore, memory systems that facilitate continuous memory management are essential for achieving long-term conversational intelligence~\cite{wang2025mirix, ouyang2025reasoningbank}. 

Existing memory systems for dialogue agents have alleviated this problem to some extent, such as those based on Retrieval-Augmented Generation (RAG)~\cite{lewis2020retrieval,jimenez2024hipporag, hu2025memory} and operating system-inspired systems like MemOS~\cite{li2025memos}. However, most of them store information as fragmented units, which leads to incoherence in long‑term interactions. Moving forward, cognitively-inspired approaches emerge as a promising direction, exemplified by recent works like A-Mem~\cite{xu2025mem} and Nemori~\cite{nan2025nemori}. While these methods draw upon cognitive concepts such as episode segmentation or short- and long-term memory distinctions, their underlying memories often lack dynamic reorganization—a process central to human memory consolidation.

\begin{figure}[t]
\centerline{\includegraphics[width=0.48\textwidth,height=0.24\textwidth]{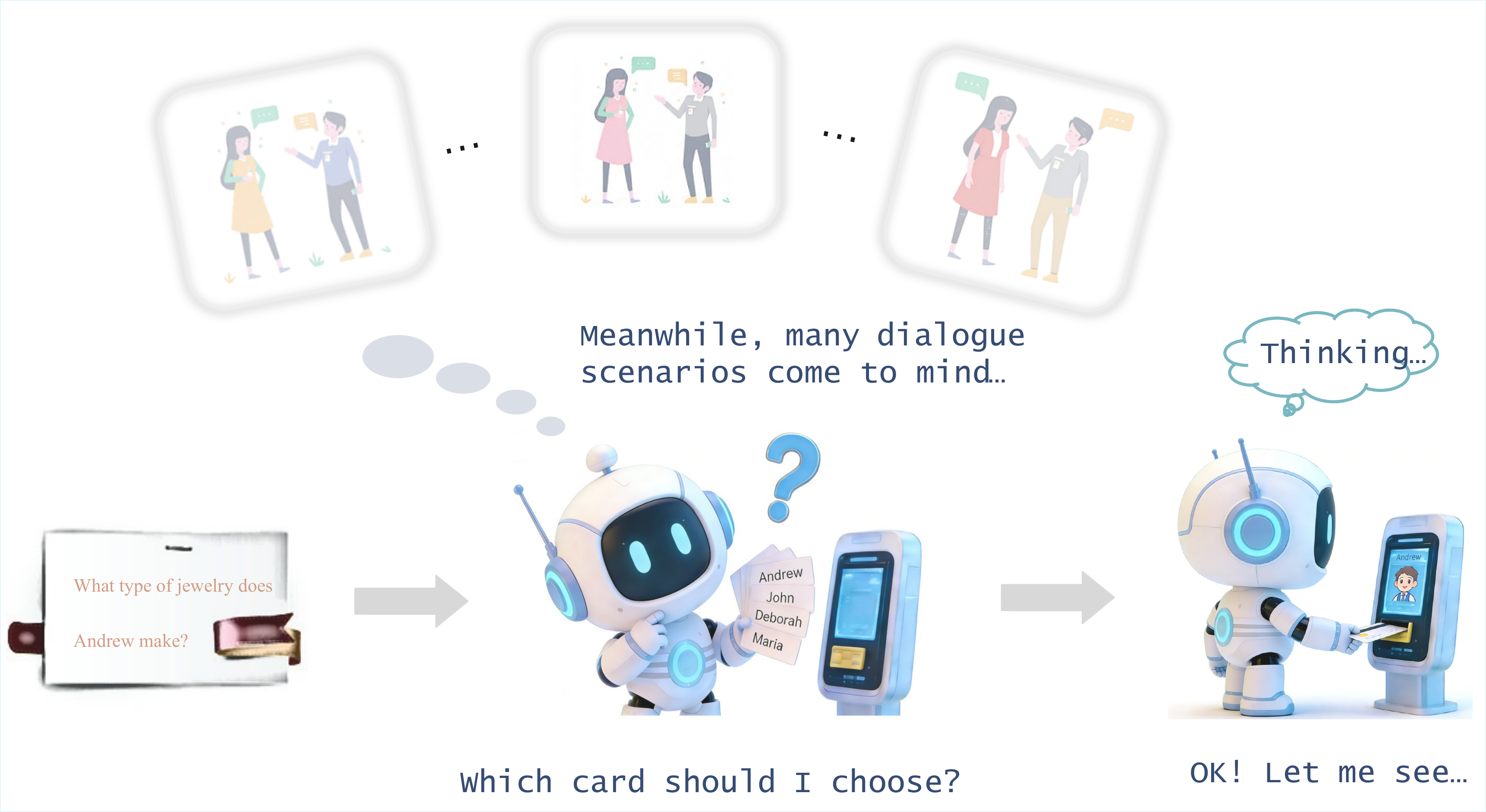}}
\caption{"Remembering" of Tracemem paradigm mimics the human cognitive process of memory recall. When queried about an individual, humans evoke a coherent personal impression while simultaneously tracing back to the specific episodes from which that knowledge was acquired.}
\label{demo}
\end{figure}

\begin{figure*}[t]
  \centering
  \includegraphics[scale=0.215,trim=0 -2 0 0, clip]{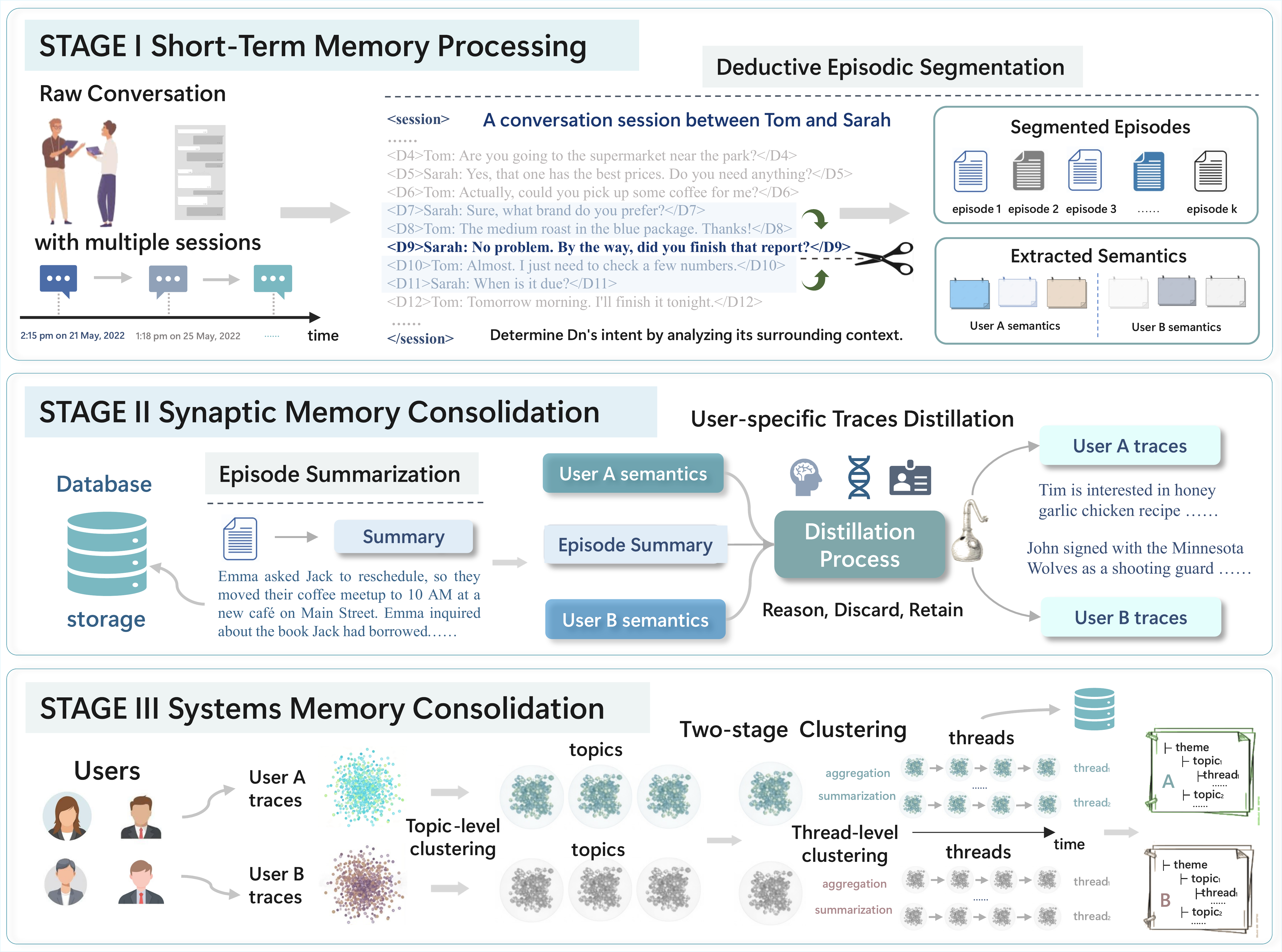}
  \caption{TraceMem Memory Construction Pipeline. TraceMem transforms raw dialogues into long-term patterns through three progressive stages. Initially, short-term memory processing partitions interaction sessions into episodes and extracts semantic representations through structured XML-based prompting. Subsequently, synaptic memory consolidation facilitates the refinement of information, distilling specific user traces from these episodic segments via a process of summarization and experience distillation. Finally, systems memory consolidation orchestrates the long-term organization through a two-stage clustering mechanism, ultimately encapsulating them into persistent, structured user memory cards.}
  \label{Outlilne}
\end{figure*}

Memory consolidation involves two principal mechanisms: the synaptic consolidation, which rapidly stabilizes recent memory traces, and the systems consolidation, a process that gradually reorganizes memory representations into enduring, long-term forms~\cite{frankland2005organization, ko2025systems, dudai2015consolidation}. Traditionally, long-term memory has been categorized into two primary systems: episodic memory, which records specific events., and semantic memory, which stores general world knowledge. Renoult et al.~\cite{renoult2012personal} conceptualized personal semantics as residing at the crossroads of episodic and semantic memories. They posited it as a heterogeneous construct encompassing autobiographical facts, repeated events, self-knowledge, and personally important concepts, which thus cannot be simply categorized as either episodic or semantic memories.

Inspired by the theory of memory consolidation and personal semantics, we propose TraceMem, a cognitively grounded memory system that weaves long-term conversational contexts into coherent narrative memory schemata, as shown in Fig.~\ref{Outlilne}. The core objective is to orchestrate fragmented conversational histories into a structured, self-evolving long-term memory that supports proactive agentic retrieval. The framework consists of two primary components: memory construction and agentic search. The construction process operates through a three-stage procedure: First, in short-term memory processing, the system partitions the continuous dialogue stream into discrete episodes by detecting topic shifts and extracting their corresponding semantic representations. Building upon this, synaptic memory consolidation synthesizes each episode into a structured episodic memory. From these, user-specific personal experience traces are distilled by aligning episodic contexts with underlying semantic knowledge. In the system consolidation stage, we cluster dispersed experience traces into hierarchical narrative threads. These threads are then encapsulated into structured memory cards, yielding dynamic and coherent narrative memory schemata for long-term conversational intelligence. Agentic Search mimics the human cognitive ability to maintain a coherent persona of an individual while retaining the capacity to recall the specific episodes from which that knowledge was acquired. To this end, the system concurrently retrieves episodic memories and strategically selects user memory cards to retrieve the narrative threads required for agentic reasoning, as shown in Fig.~\ref{demo}. In summary, our key contributions are threefold:
\begin{itemize}[leftmargin=*, nosep]
    \item \textbf{Cognitive-Inspired Framework:} We propose TraceMem, a cognitively grounded memory system that transforms fragmented dialogue histories into coherent, self-evolving narrative memory schemata.
    \item \textbf{Memory Consolidation and Agentic Search:} We develop a memory consolidation mechanism including synaptic and systems consolidation to enable a persistent persona. Agentic search enables human-like source-attribution, empowering the agent with more precise reasoning.
    \item \textbf{Empirical Validation:} We conduct extensive experiments across different backbones, demonstrating that TraceMem significantly outperforms state-of-the-art memory systems in both retrieval accuracy and complex reasoning.
\end{itemize}

\section{Related Works}
\subsection{From Long Context to Memory Systems}
While various techniques extend context windows of Large Language Models (LLMs), raw capacity does not equate to effective utilization. Mere context extension remains an incomplete solution for true memory persistency, as it is plagued by inherent utilization bottlenecks like the “Lost-in-the-Middle” phenomenon~\cite{liu2024lost, hu2026evermemos, zhang2025lost}, This suggests that as the input space grows, the model's ability to selectively recall and reason over distributed information actually diminishes, necessitating a shift from passive expansion to active memory management. Accordingly, there are several existing categories of memory systems. Mem0~\cite{chhikara2025mem0} coordinates a two-stage process to synchronize salient facts within vector storage, utilizing a graph-based alternative to refine the persistence and retrieval of long-term history. Hierarchical and structured approaches like H-MEM~\cite{sun2025hierarchical} and G-Memory~\cite{zhang2025g} organize data into multi-level structures to improve retrieval quality. To incorporate temporal-aware capabilities, Zep~\cite{rasmussen2025zep} uses its Graphiti engine to dynamically track relationships among entities. TiMem~\cite{li2026timem} organizes interaction histories into hierarchical representations across increasing temporal scales. MemoTime~\cite{tan2025memotime} enforces monotonic timestamp constraints by decomposing complex queries into a Tree of Time. Another direction is operating system-inspired management, including frameworks like MemGPT~\cite{packer2023memgpt}, MemoryOS~\cite{kang2025memory}, and MemOS~\cite{li2025memos}. However, despite these advancements in memory management, existing systems still fail to establish a well-defined user persona from accumulated interactions. More critically, they often lack the effective utilization of agentic search to actively navigate and synthesize complex histories.

\subsection{Cognitive-Inspired Memory Architectures}
Current research is gradually moving beyond simple passive storage mechanisms by introducing more advanced cognitively inspired memory models that draw upon cognitive theories to organize information more effectively. For instance, A-Mem~\cite{xu2025mem} draws on the core idea of the Zettelkasten note-taking method~\cite{kadavy2021digital}, organizing fragmented memory entries into dynamic, interconnected knowledge networks. Nemori~\cite{nan2025nemori}, grounded in Event Segmentation Theory~\cite{zacks2001event, zacks2007event}, employs a predict-calibrate mechanism to segment dialogue streams into semantically bounded event episodes. LightMem~\cite{fang2025lightmem}, inspired by the Atkinson-Shiffrin human memory model~\cite{atkinson1968human}, facilitates the gradual transition of information from transient perception to long-term storage through lightweight sensory filtering and offline sleep-time updates.
Current models tend to archive user experiences as discrete and independent episodes, lacking the necessary structural glue to interlink them. This fragmentation prevents the system from synthesizing isolated interactions into a coherent, diachronic body of user-specific cognition, thereby limiting its ability to grasp the broader context of a user's life evolution. TraceMem addresses this gap by systematically organizing disjointed interaction traces into evolving narrative threads that represent the user's ongoing life story. 



\section{TraceMem}
This section outlines the architectural workflow, covering the progression from short-term memory processing to synaptic and systems consolidation, followed by a description of the agentic search mechanism.


\subsection{Short-Term Memory Processing}




Short-term memory is the capacity for holding a small amount of information in an active, readily available state for a short interval. In our framework, short-term memory serves as the session processing buffer, where raw conversational input is analyzed in real time for topic segmentation and semantic representation generation.
\subsubsection{\bfseries Deductive Episodic Segmentation.}
Our system employs a deductive reasoning approach for dialogue episode segmentation, implemented using XML-based structured prompts. The method classifies each utterance into one of two intent categories: \textit{topic change} (TC), indicating the introduction of a new subject, activity, or domain shift, and \textit{topic development} (TD), referring to direct responses or elaborations on the current topic without introducing new content. This classification maintains conversational coherence while preventing excessive topic accumulation.

Formally, for each dialogue utterance $D_n$ in a session, the system analyzes its bidirectional relationships with preceding context $D_{pre}$ and subsequent context $D_{sub}$ to infer discourse intent. This process is defined as a mapping function $\mathcal{F}$ that determines the intent $I_n$:
\begin{equation}
I_n = \mathcal{F}(D_n, \mathcal{D}_{pre}, D_{sub}) = 
\begin{cases} 
\text{TC}, & \text{if } \mathcal{G}(D_n, \mathcal{D}_{pre}, D_{sub})\\
\text{TD}, & \text{otherwise}
\end{cases}
\end{equation}
where $\mathcal{G}(D_n, \mathcal{C}_{pre}, D_{sub})$ evaluates whether the utterance $D_n$ indicates a topic shift. This classification mechanism maintains conversational coherence while enforcing structured boundaries.

Thus, the system partitions sessions in a conversation into a sequence of episodes $\mathcal{M}_{\text{epi}}=\{E_1, E_2, \dots, E_k\}$, where $k$ represents the number of segments. Each episode $E_i$ is a contiguous subsequence of utterances sharing the same topic.

\subsubsection{\bfseries Semantic Memory Extraction.}
During short-term memory processing, the system generates semantic representations in real-time to extract and structure key information from the dialogue flow. These transient memories provide a structured content base for subsequent memory consolidation.

\begin{figure}[t]
    \centering
    \begin{subfigure}[b]{0.46\textwidth}
        \centering
        \includegraphics[width=\textwidth]{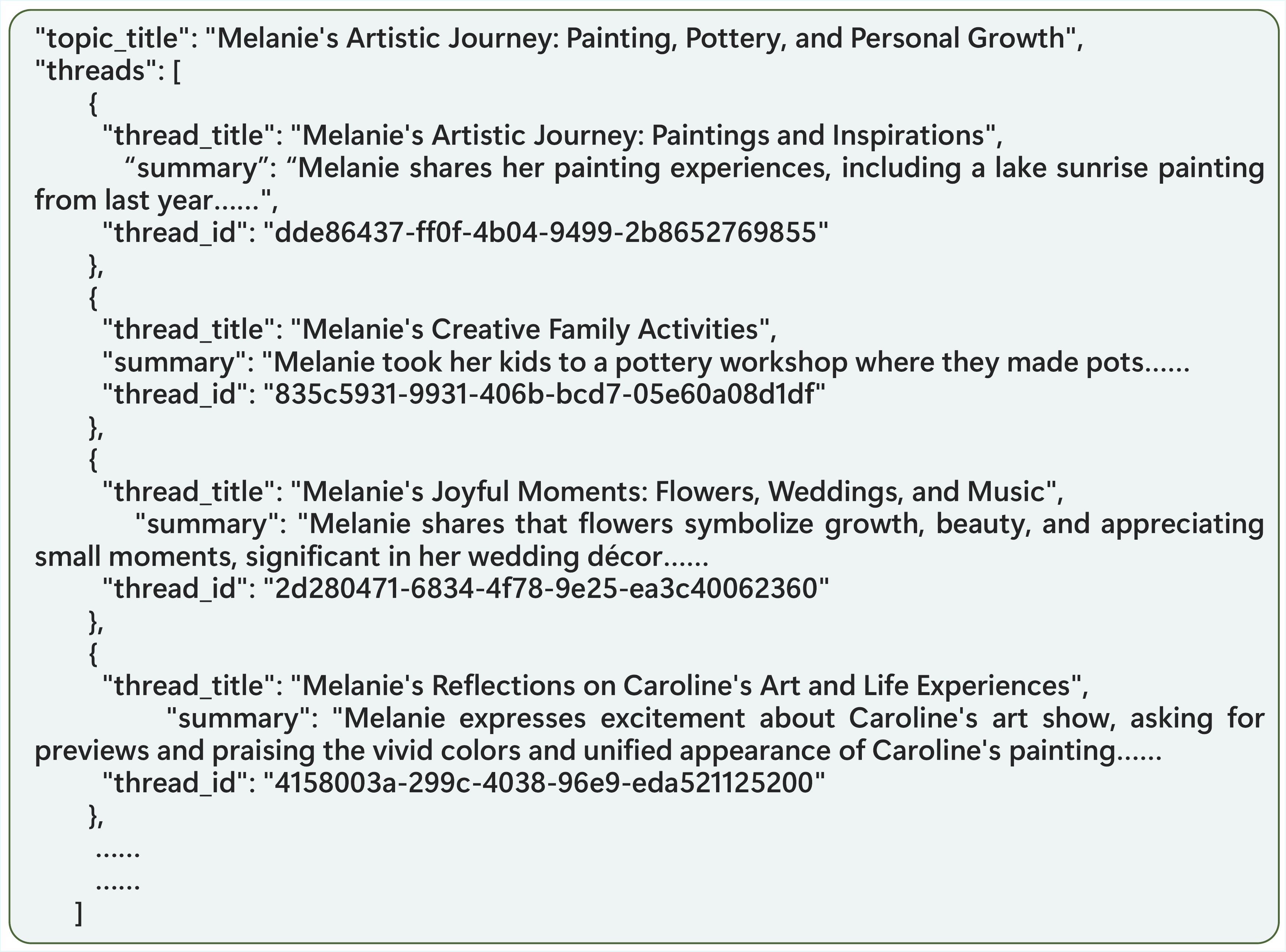}
        \caption{A topic in Melanie's Memory Card (Melanie's Journey of Self-Care, Family Activities, and Community Involvement)}
        \label{fig:card_a}
    \end{subfigure}

    \begin{subfigure}[b]{0.46\textwidth}
        \centering
        \includegraphics[width=\textwidth]{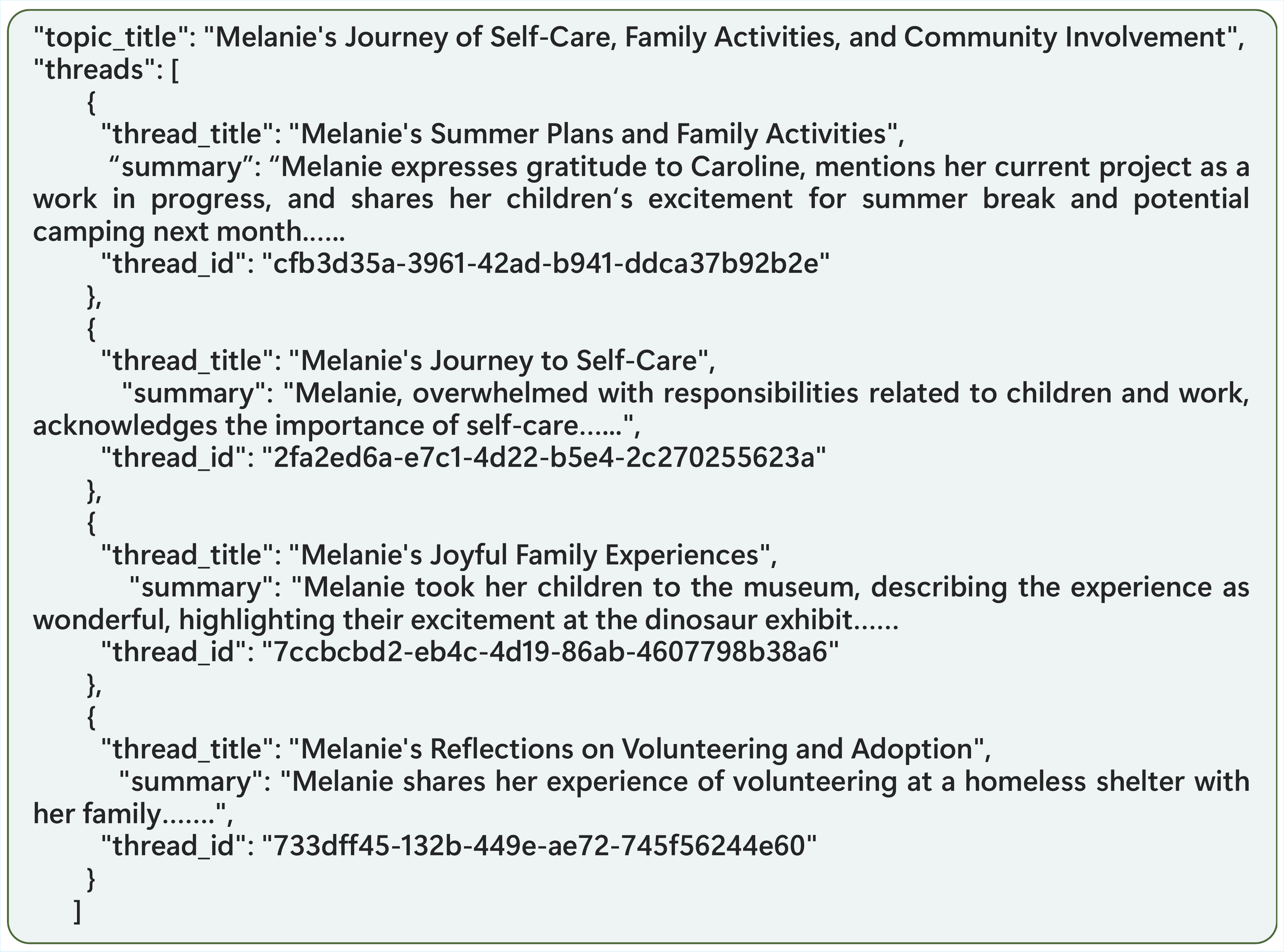} 
        \caption{A topic in Melanie's Memory Card (Melanie's Artistic Journey: Painting, Pottery, and Personal Growth)}
        \label{fig:card_b}
    \end{subfigure}
    
    \caption{Examples of topics in a user memory card.}
    \label{fig:card}
\end{figure}

Upon segmenting the topic, the system extracts structured factual information to form transient semantics. Let $s_n$ be the structured semantics extracted from utterance $D_n$. The extraction process is formulated as a transformation $\mathcal{T}$ that maps raw dialogue and associated visual metadata into a fact-based space:
\begin{equation}
s_n = \mathcal{T}(D_n \mid D_{n-1}, D_{n-2}, \text{Img}_n)
\end{equation}
where $s_n$ is defined as a union of exhaustive factual elements derived from both textual and visual inputs. The system then generates a corresponding semantic set sequence $\mathcal{M}_{\text{sem}} = \{S_1, S_2, \dots, S_k\}$, each $S_i = \{s_1, s_2, \dots, s_m\}$ represents a set of semantics whose source is the $i$-th episode $E_i$.

The segmented episodes $\mathcal{M}_{\text{epi}}$ and semantic set $\mathcal{M}_{\text{sem}}$ forms the output of the short‑term memory module, providing foundation for subsequent long‑term memory consolidation.

\subsection{Synaptic Memory Consolidation}
Our system formally models this process through two sequential: episode summarization and user experience distillation. 

For each conversational episode \( E_i \), the system will generates a episodic summary $E_{s,i}$. Subsequently, user experience distillation distills experience traces by jointly processing the episode summaries and the semantic set \( \mathcal{M}_{\text{sem}} \). It applies a rule‑based filter \( \mathcal{R} \) defined in the prompt to extract concrete biographical facts. The output is a user experience trace \( ET_i \), defined by the function \( \Phi \): \( ET_i = \Phi(E_{s,i}, S_i; \mathcal{R}) \). The filter $\mathcal{R}$ retains only the users' own personal experiences and contextual details.

The Synaptic Memory Consolidation (Syna-MC) process is formulated as a transformation that maps episodes and their associated semantic memories into a dual-output: a set of condensed episode summaries and a sequence of experience traces:
\begin{equation}
\text{Syna-MC} \bigl(\mathcal{M}_{\text{epi}}, \mathcal{M}_{\text{sem}}\bigr) \rightarrow \bigl(\mathcal{M}_{\text{epi,s}}, U\bigr),
\end{equation}
where $\mathcal{M}_{\text{epi,s}} = \{E_{s,1}, E_{s,2}, \dots, E_{s,k}\}$ is the set of episode summaries. $U = \{ET_1, ET_2, \dots, ET_z\}$ is the set of personal experience traces distilled by the function $\Phi$, where $z$ represents the total number of identified traces in a conversation.

\subsection{Systems Memory Consolidation}
In our framework, systems memory consolidation translates to the transformation of the experience-level traces obtained from synaptic memory consolidation into persistent, organized user memory cards. This process operates on two interconnected levels: (1) Traces Clustering and Narrative Thread Formation that uncovers latent narrative threads, and (2) Constructing the User Narrative Schemata that encapsulates these threads into a coherent story under topics.

\subsubsection{\bfseries Traces Clustering and Narrative Thread Formation}
We employ a two‑stage clustering pipeline. First, PCA~\cite{abdi2010principal} reduces the dimensionality of the high‑dimensional embeddings to retain essential variance. The reduced vectors are then processed by UMAP~\cite{mcinnes2018umap}, which learns a low‑dimensional manifold that preserves both local and global semantic relationships. Then HDBSCAN~\cite{mcinnes2017hdbscan} performs density‑based clustering on this manifold to identify core groups while flagging low‑density points as noise. Finally, a KNN~\cite{guo2003knn} classifier reassigns these noise points to the nearest coherent cluster.

\begin{figure}[t]
\centerline{\includegraphics[width=0.48\textwidth,height=0.22\textwidth]{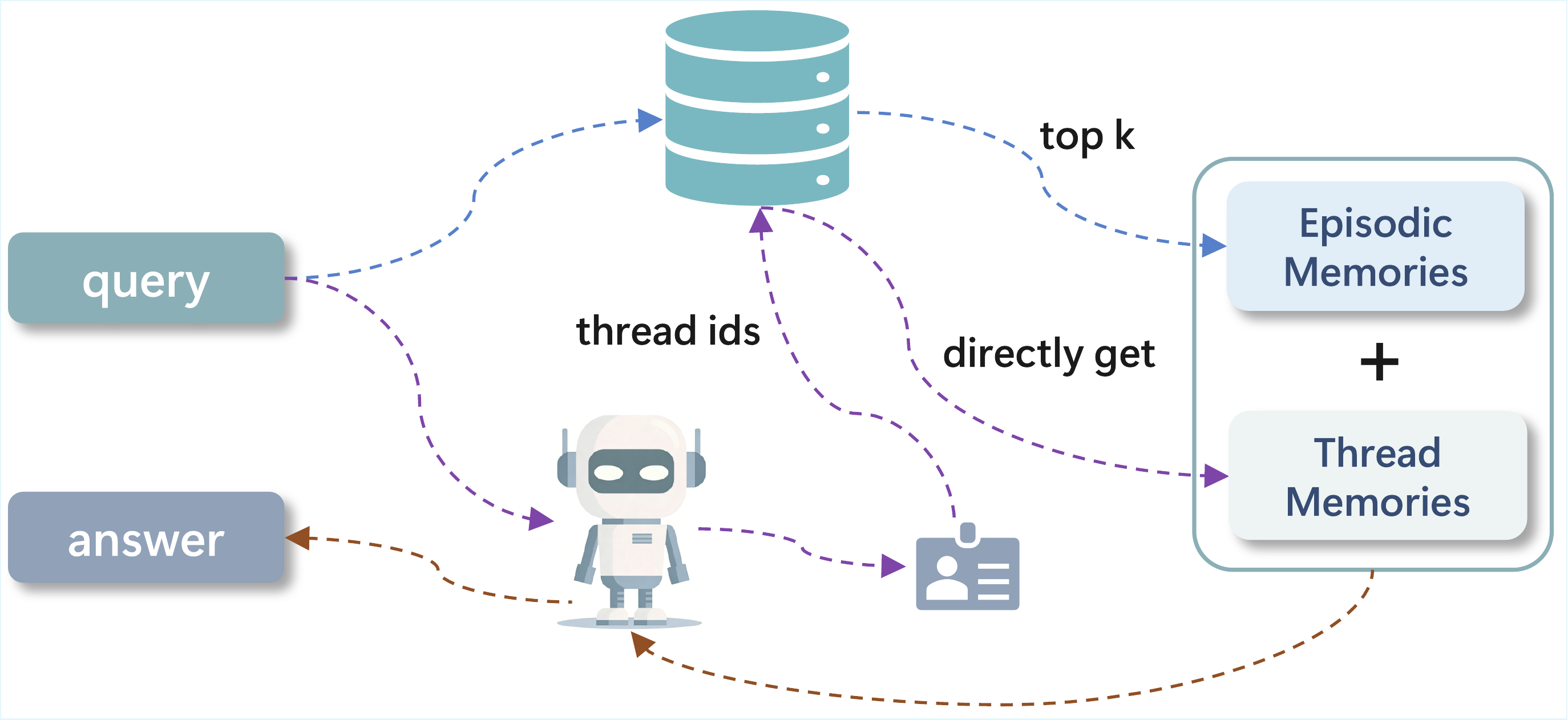}}
\caption{Agentic Search. The system retrieves top-$k$ episodic memories (blue line) while searching the user card to identify and directly fetch relevant thread traces via their IDs (purple line). This retrieved information is then integrated into the reasoning engine to generate historically consistent responses (red line).}
\label{search}
\end{figure}

The atomic units for system consolidation are the experience traces $U$ generated during synaptic consolidation. We employ a two-stage hierarchical clustering strategy:

\noindent \textbf{Topic-level (coarse) clustering}: It groups the entire set of experience traces into distinct topic clusters based on their semantic similarity. Formally, given the set of experience embeddings $\mathcal{E} = \{\mathbf{e}_1, \mathbf{e}_2, \dots, \mathbf{e}_k\}$, the coarse clustering operation $\mathcal{C}_{\text{topic}}$ produces a partition into $m$ topic clusters:
\begin{equation}
    \{T_1, T_2, \dots, T_c\} = \mathcal{C}_{\text{topic}}(\mathcal{E},\theta_{\text{topic}}),
\end{equation}
where each cluster $T_i$ corresponds to a macro‑topic. where \( \theta_{\text{topic}} \) denotes the clustering parameters.

\noindent \textbf{Thread‑level (fine‑grained) clustering}: It organizes the temporally ordered experiences within each topic cluster into coherent narrative threads based on their fine‑grained semantic similarity. Formally, for each topic cluster $T_i$, the fine‑grained clustering operation \( \mathcal{C}_{\text{thread}} \) yields a set of threads:
\begin{equation}
\{t_{i1}, t_{i2}, \dots, t_{if}\} = \mathcal{C}_{\text{thread}}(T_i; \theta_{\text{thread}}),
\end{equation}
Each resulting thread \( t_{ij} \) aggregates a subset of experiences from \( T_i \) that are both semantically proximate and chronologically sequential, thereby forming a cohesive narrative unit under the broader topic \( T_i \). This two‑stage hierarchical clustering thus transforms a temporal stream of experiences into a structured set of narrative threads, ready for encapsulation into higher‑level memory schemata.

\subsubsection{\bfseries Constructing the User Narrative Schemata}
The construction of memory cards organizes the thematically clustered narrative threads into a structured knowledge representation with clear semantic hierarchy. This process integrates discrete narrative units into a coherent memory schema, manifesting as a three‑level encapsulation: a top‑level theme title that summarizes the overall narrative arc; a middle layer of topical sections, each corresponding to one coarse‑level thematic cluster; and a bottom layer of thread entries, where each narrative thread is represented by a descriptive title, a coherent summary, and a unique identifier, as shown in Fig.~\ref{fig:card}. Through this structured encapsulation, the clustering outputs are transformed into persistent and structured narrative memory schemata.

\subsection{Agentic Search}
Upon receiving a query, the Agentic Search mechanism proceeds through three sequential stages, as shown in Fig.~\ref{search}:

\begin{itemize}[leftmargin=*, nosep]
    \item \textbf{Episodic Memory Retrieval:} The query is used as a semantic key to search a global vector database of episodic memory, retrieving the top-$K$ most relevant memory fragments.

    \item \textbf{Memory Card Selection:} Based on user or topic cues identified directly from the query, the system selects one or more relevant memory cards.

    \item \textbf{Thread Extraction:} For each selected memory card, the agent examines its internal structure, including topic titles and the associated thread entries each comprising a title, summary, and unique thread ID, to pinpoint relevant narrative threads. The corresponding thread content is then retrieved directly from the vector database.
\end{itemize}

The system subsequently integrates the discrete episodic memories with the personal narrative threads extracted from the selected cards. This fusion creates a multi-dimensional contextual foundation, enabling coherent reasoning and response generation.

\section{Experiement}
This section details the experimental evaluation of TraceMem. We first describe the setup, followed by a presentation of our primary results and subsequent ablation studies.

\subsection{Experiement Setup}

\subsubsection{\bfseries Benchmark Dataset} We validate our memory system on LoCoMo benchmark dataset~\cite{maharana2024evaluating}. The dataset comprises 10 long dialogues with an average of 600 turns per dialogue. It includes 1,540 annotated questions spanning four reasoning categories: single-hop, multi-hop, open-domain, and temporal reasoning. Multi-hop reasoning: Questions that require synthesizing information from multiple, distinct dialogue turns or sessions. Temporal reasoning: Questions that necessitate understanding temporal cues and event sequences within the dialogue history. Open-domain reasoning: Questions that demand integration of dialogue content with external commonsense. Single-hop reasoning: Questions answerable by retrieving information from a single dialogue turn.

\begin{table}[h]
    \centering
    \renewcommand{\arraystretch}{1.2} 
    \setlength{\tabcolsep}{3pt}    
      \caption{Key Characteristics of Evaluated Methods.}
    \begin{tabular}{lcccc}
        \toprule
        \textbf{System Traits} & \textbf{TraceMem} & \textbf{A-Mem} & \textbf{LightMem} & \textbf{Nemori}  \\
        \midrule
        Persona Model   & \ding{51}&  \ding{55}& \ding{55} &   \ding{55}\\
        Agentic Search   & \ding{51} & \ding{55} & \ding{55} &  \ding{55}  \\
        Conginitive     & \ding{51} &  \ding{51} &  \ding{51}& \ding{51}   \\
        Episode Seg.    & \ding{51} &  \ding{55} &  \ding{51}& \ding{51}   \\
        \bottomrule
    \end{tabular}
    \label{method_traits}
\end{table}

\begin{table*}[t]
\centering
\setlength{\tabcolsep}{12pt} 
\setlength{\aboverulesep}{0pt}    
\setlength{\belowrulesep}{0pt}    
\renewcommand{\arraystretch}{1.1}
\caption{Main Comparison Results on Different Backbones.}
\label{tab:main_results}
\begin{tabular}{c | l | c | c  |c  |c |c}
\toprule
\textbf{Backbone} & \textbf{Method}  & \textbf{SingleHop} & \textbf{MultiHop} & \textbf{Temporal} & \textbf{OpenDomain} & \textbf{Overall Acc.} \\ \midrule
\multirow{6}{*}{\rotatebox{90}{\makebox[0pt]{GPT-4o-mini}}} 
& FullText  &0.8395 &0.6986  & 0.5171 & 0.5833 &  0.7305\\
 & NaiveRAG  & 0.5660 & 0.3369 & 0.3084 & 0.4375 & 0.4623 \\
 & A-Mem     & 0.4792 & 0.3298 & 0.3832 & 0.1771 & 0.4130\\
  & LightMem  & 0.7414 & 0.6069 & 0.7422 & 0.4819 & 0.7007 \\
 & Nemori   &  0.8252& 0.6312 & 0.6854& 0.5000 & 0.7403 \\
 & \textbf{TraceMem} & \textbf{0.9310} & \textbf{0.9220} & \textbf{0.8660} & \textbf{0.7083} & \textbf{0.9019}  \\ \midrule
\multirow{5}{*}{\rotatebox{90}{\makebox[0pt]{GPT-4.1-mini}}} 
& FullText  &0.9203 & 0.8723 & 0.8162 & 0.6667 & 0.8740 \\
 & NaiveRAG & 0.6207 & 0.3901 & 0.3769 & 0.4792 & 0.5188 \\
 & A-Mem     & 0.6147 & 0.5603 & 0.6355 & 0.3958 & 0.5955 \\
 & LightMem  &  0.8121 & 0.7057 & 0.7882 & 0.4792 & 0.7669 \\
 & Nemori    & 0.8704 & 0.7482 & 0.7632 & 0.5417 & 0.8052 \\
 & \textbf{TraceMem}  & \textbf{0.9512} & \textbf{0.8936} & \textbf{0.9097} & \textbf{0.8021} & \textbf{0.9227} \\ \bottomrule
\end{tabular}
\end{table*}

\subsubsection{\bfseries Baselines} To assess the performance of our proposed memory system, TraceMem, a comprehensive evaluation is conducted by comparing it against a range of baseline approaches. The following is a list of all the baseline methods used in the comparison:

\begin{itemize}[leftmargin=*, nosep]
    \item \textbf{FullText} Full Context provides the entire dialogue history to the LLM, serving as the theoretical upper bound for information availability.
    \item \textbf{NaiveRAG} NaiveRAG is a standard RAG implemention  where dialogues are first chunked into fixed-length segments and retrieve relevant segments to augment generation.
    \item \textbf{A-mem}~\cite{xu2025mem} A-Mem organizes memories into structured notes, establishes connections between related memories, drawing inspiration from the Zettelkasten method.
    \item \textbf{LightMem}~\cite{fang2025lightmem} LightMem is a lightweight memory system for LLMs, inspired by the Atkinson-Shiffrin human memory model.
    \item \textbf{Nemori}~\cite{nan2025nemori} Nemori is a cognitively-inspired memory system that incorporates the Two-Step Alignment Principle and the Predict-Calibrate Principle.
\end{itemize}

Table~\ref{method_traits} compares the configurations of four memory management methods across key system traits. Specifically, TraceMem is the only system that supports both the persona model and agentic search. Inspiration from cognitive principles is consistent across all systems. TraceMem, LightMem, and Nemori implement episodic segmentation, whereas A-Mem lacks this mechanism.

\subsubsection{\bfseries Evaluation Scheme} We use accuracy as the evaluation metric to assess whether model outputs semantically match the ground truth. Our evaluation utilizes GPT-4o-mini as an LLM judge, guided by the prompt template provided by Nemori~\cite{nan2025nemori}. To ensure a fair and consistent comparison, all models are evaluated under the same judge prompt. Detailed specifications of the judge prompt is documented in the Appendix.
\subsubsection{\bfseries Implemetation Details} We use GPT-4o-mini and GPT-4.1-mini as the backbone language models. Text embeddings are generated with the \texttt{text-embedding-3-small} model and indexed in a ChromaDB~\cite{chromadb2022} vector database for similarity search. Our hierarchical clustering approach combines UMAP for dimensionality reduction and HDBSCAN for clustering. We define \(\theta_{\mathrm{nei}}\) and \(\theta_{\mathrm{mc}}\) as the key parameters for UMAP (\texttt{n\_neighbors}) and HDBSCAN (\texttt{min\_cluster\_size}), respectively. The parameters are set to \(\theta_{\mathrm{nei}}=10, \theta_{\mathrm{mc}}=5\) for topic-level clustering ($\theta_{\text{topic}}$) and \(\theta_{\mathrm{nei}}=2, \theta_{\mathrm{mc}}=2\) for thread-level clustering ($\theta_{\text{thread}}$). For episodic memory retrieval, the top $K=10$ most relevant memories are selected.


\subsection{Primary Results}
The proposed framework was evaluated across two backbones: GPT-4o-mini and GPT-4.1-mini. As illustrated in Table~\ref{tab:main_results}, TraceMem achieves a decisive lead in overall accuracy across both settings. On the GPT-4o-mini backbone, TraceMem attains an overall score of $0.9019$, outperforming the strongest baseline, Nemori, by approximately 16\%. This performance exceeds the FullText upper bound by roughly 17\%. This dominance is consistent on the GPT-4.1-mini, where TraceMem’s overall score rises to 0.9227, maintaining about 30\% lead over A-mem and 16\% lead over LightMem. These results demonstrate that TraceMem delivers substantial and stable performance gains across different backbone models. Furthermore, TraceMem exhibits consistent performance across all four task categories, with a particular capacity for integrating long-term information in MultiHop and Temporal reasoning tasks.

TraceMem demonstrates marked superiority in Temporal tasks requiring chronological reasoning. On the GPT-4o-mini backbone, TraceMem achieves a high score of 0.8660, surpassing the best-performing baseline LightMem (+12\%). On the GPT-4.1-mini backbone, performance reaches 0.9097, outperforming the best baseline FullText by approximately 10\%. In sharp contrast, the traditional NaiveRAG struggles significantly, with TraceMem outperforming it by more than 50\% across both backbones. These results indicate that the design of TraceMem facilitates more precise temporal association, thereby effectively addressing the challenges of complex time-sensitive reasoning. In MultiHop tasks, TraceMem exhibits a consistent performance advantage. On GPT-4o-mini, TraceMem achieves a score of 0.9220, outperforming the FullText baseline by about 20.0\%. On the GPT-4.1-mini backbone, its score of 0.8936 firmly holds the first place. Notably, while almost all baseline methods suffer a significant drop in MultiHop tasks compared to their SingleHop results, TraceMem consistently maintains its accuracy near or above the 0.90 threshold. Compared to NaiveRAG, TraceMem achieves a 60\% improvement (0.3369 on GPT-4o-mini and 0.3901 on GPT-4.1-mini), demonstrating enhanced performance in multi-step inference and information integration.


\begin{table}[h]
    \centering
    \renewcommand{\arraystretch}{1.2} 
    \setlength{\tabcolsep}{3pt}    
    \caption{Statistical Summary of Processed Memory Components of TraceMem}
    \begin{tabular}{lccccc}
        \toprule
        \textbf{Backbone} &   \textbf{Episode} & \textbf{Exper.} & \textbf{Thread}  & \textbf{Discard} & \textbf{Dis. Rate}\\
        \midrule
        GPT-4o-mini   &925   & 1379 &   475 &235.5 &25.46\%\\
        GPT-4.1-mini      & 1212 & 2104 &  736  &160& 13.20\%\\
        \bottomrule
    \end{tabular}
    \label{tab:number}
\end{table}

\begin{table*}[t]
\centering
\setlength{\tabcolsep}{7.5pt} 
\setlength{\aboverulesep}{0pt}    
\setlength{\belowrulesep}{0pt}    
\renewcommand{\arraystretch}{1.2}
\caption{Ablation Study Results.}
\label{tab:ablation_results}
\begin{tabular}{  l | c | c  |c  |c |c}
\toprule
\textbf{Method}  & \textbf{SingleHop} & \textbf{MultiHop} & \textbf{Temporal} & \textbf{OpenDomain} & \textbf{Overall} \\ \midrule
Episodic Memory Only  &0.8954  & 0.8262 & 0.7664 &  0.6875& 0.8429 \\
 Episodic Memory + Semantic Representation (20) & 0.9298&  0.8617&  0.7726& 0.6979 &  0.8701 \\
Episodic Memory + Semantic Representation (40) & \textbf{0.9334} & 0.8901& 0.7850 &0.6875 & 0.8792\\
Without Agentic Search  &0.9287  &  0.8652&  0.8349& \textbf{0.7188} &  0.8844\\
Complete TraceMem  & 0.9310 & \textbf{0.9220} & \textbf{0.8660} & 0.7083 & \textbf{0.9019}   \\ \bottomrule
\end{tabular}
\end{table*}

Table~\ref{tab:number} presents the cumulative counts, aggregated across all conversations in the dataset, for two backbone models at different stages of memory construction. In this table, Episode refers to the total number of event units shared by both users within each conversation, whereas Experience and Thread represent the sum of the corresponding memory instances from both individual users. The data shows that GPT-4.1-mini generated 1,212 shared episodes, compared to 925 for GPT-4o-mini. The higher number of episodes produced by GPT-4.1-mini indicates that it is more sensitive to topic shifts in dialogue. Experience counts total 2,104 for GPT-4.1-mini and 1,379 for GPT-4o-mini, with corresponding memory threads of 736 and 475. The number of discarded items is determined by $N_{discard} = N_{epi}-\frac{1}{2} N_{exp} $, where $N_{exp}$ and $N_{epi}$ represent the total counts of experiences and episodes, respectively. The discard rate is defined as the ratio of discarded entries to the total number of episodes per user. According to the results, GPT-4.1-mini has a discard quantity of 160 and a Discard Rate of 13.20\%, while GPT-4o-mini shows a discard quantity of 235.5 and a Discard Rate of 25.46\%. The lower discard rate of GPT-4.1-mini demonstrates a more conservative strategy during information distillation.

\begin{figure}[t]
    \centering
    \begin{subfigure}[b]{0.47\textwidth}
        \centering
        \includegraphics[width=\textwidth]{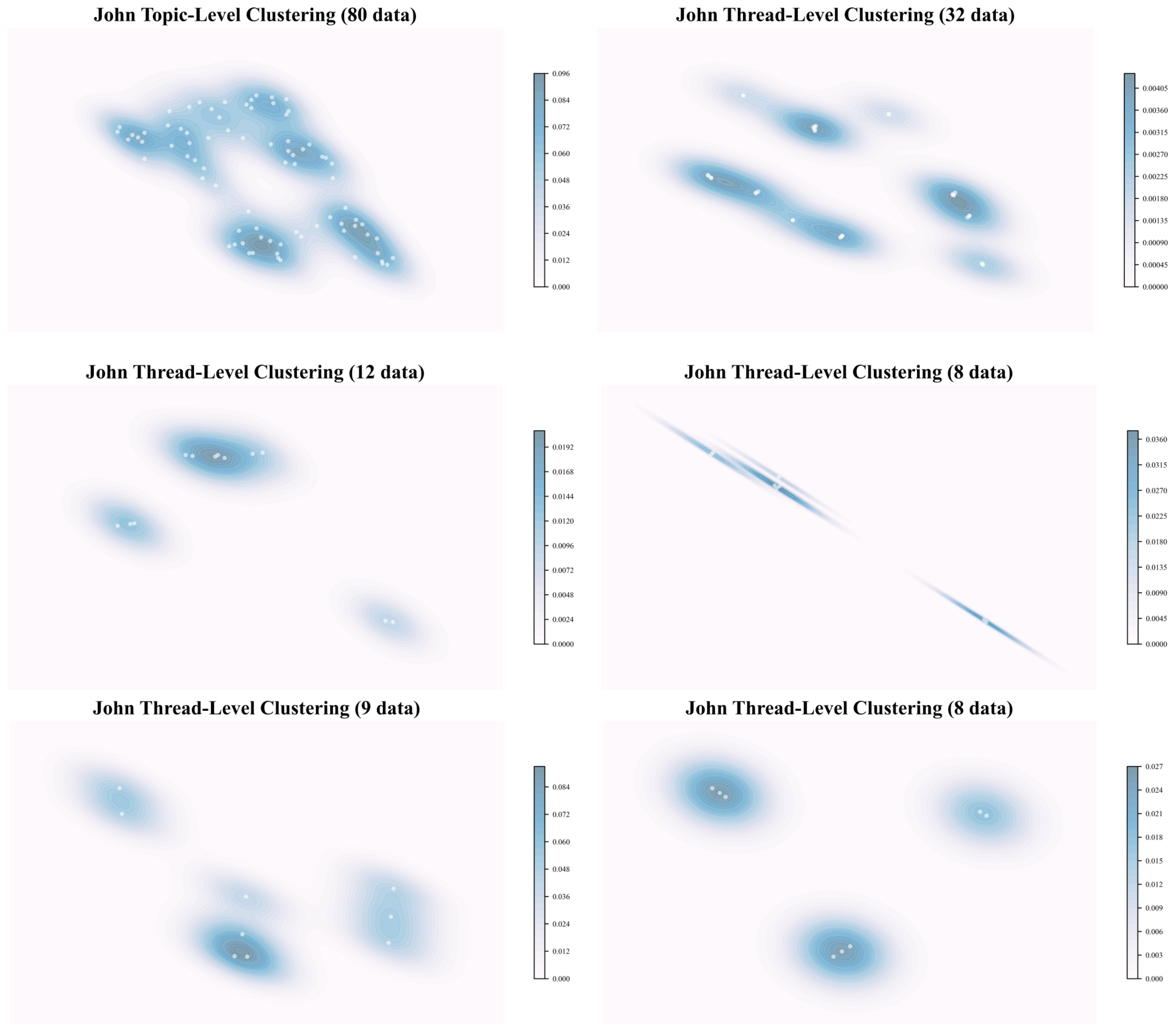}
        \caption{User A (John) Clustering Results}
        \label{fig:cluster11}
    \end{subfigure}
    \begin{subfigure}[b]{0.47\textwidth}
        \centering
        \includegraphics[width=\textwidth]{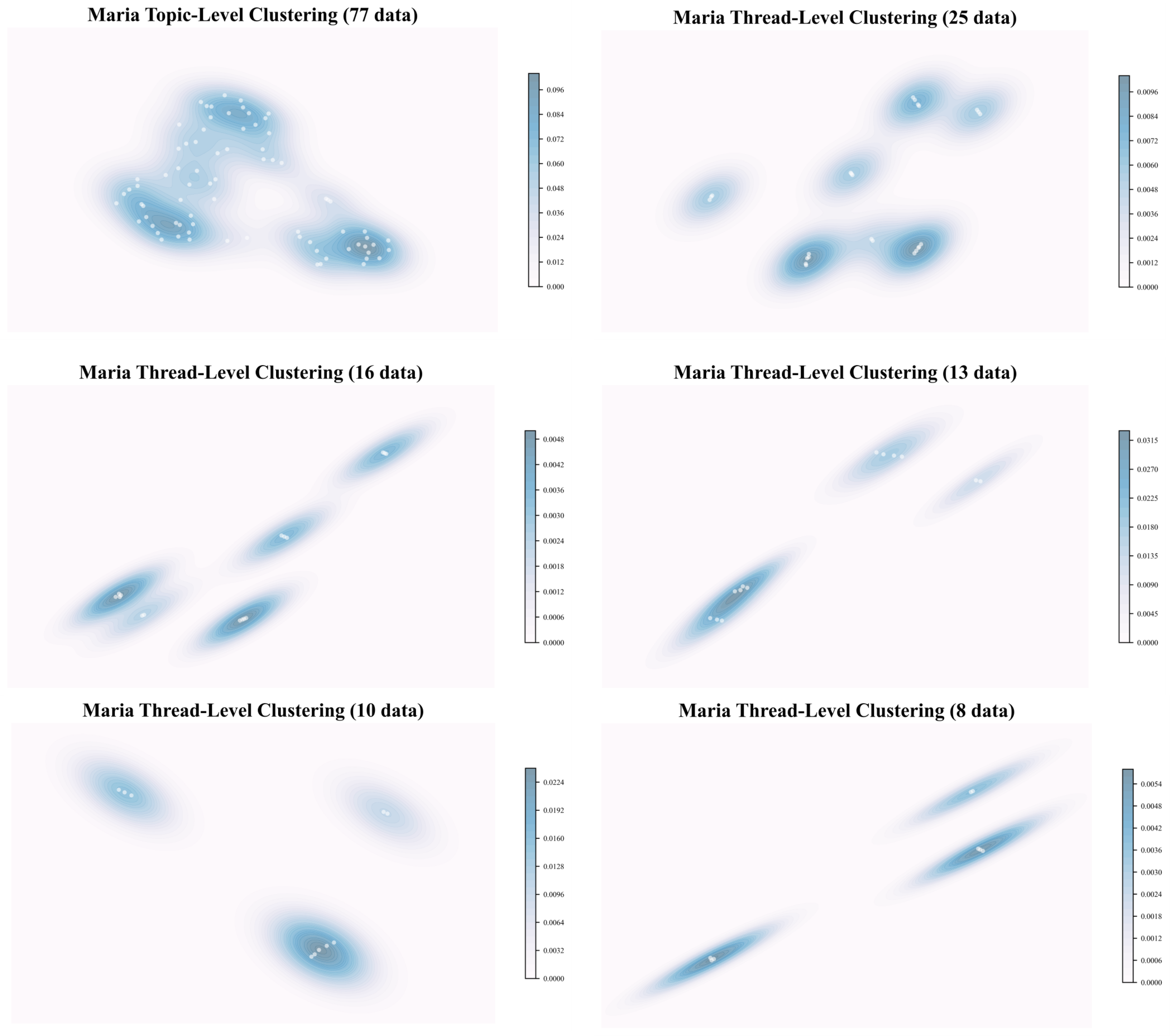} 
        \caption{User B (Maria) Clustering Results}
        \label{fig:cluster12}
    \end{subfigure}
    \caption{An example of clustering result of GPT-4o-mini backbone.}
    \label{fig:cluster1}
\end{figure}

\begin{figure}[t]
    \centering
    \begin{subfigure}[b]{0.47\textwidth}
        \centering
        \includegraphics[width=\textwidth]{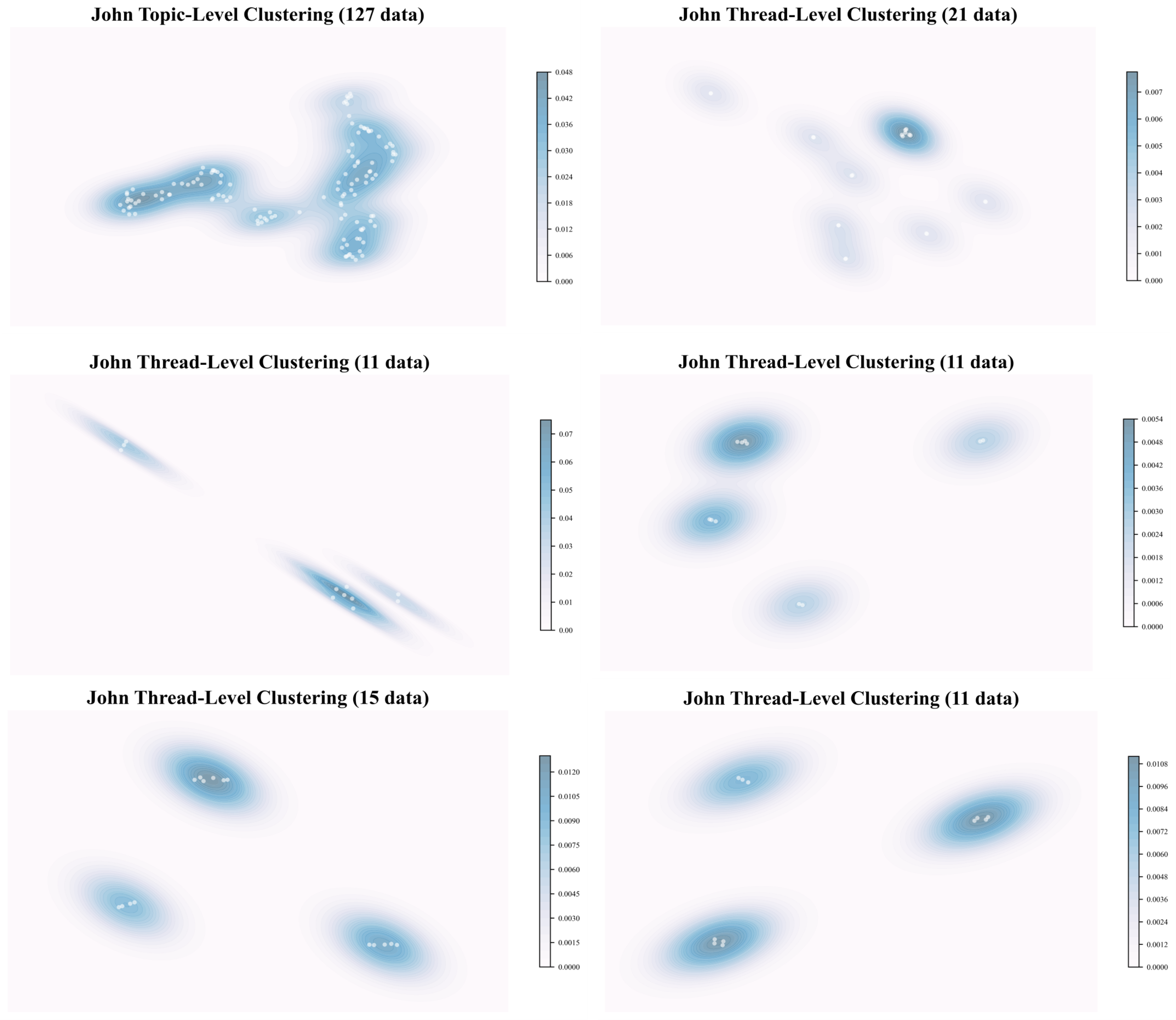}
        \caption{User A (John) Clustering Results}
        \label{fig:cluster21}
    \end{subfigure}
    \begin{subfigure}[b]{0.47\textwidth}
        \centering
        \includegraphics[width=\textwidth]{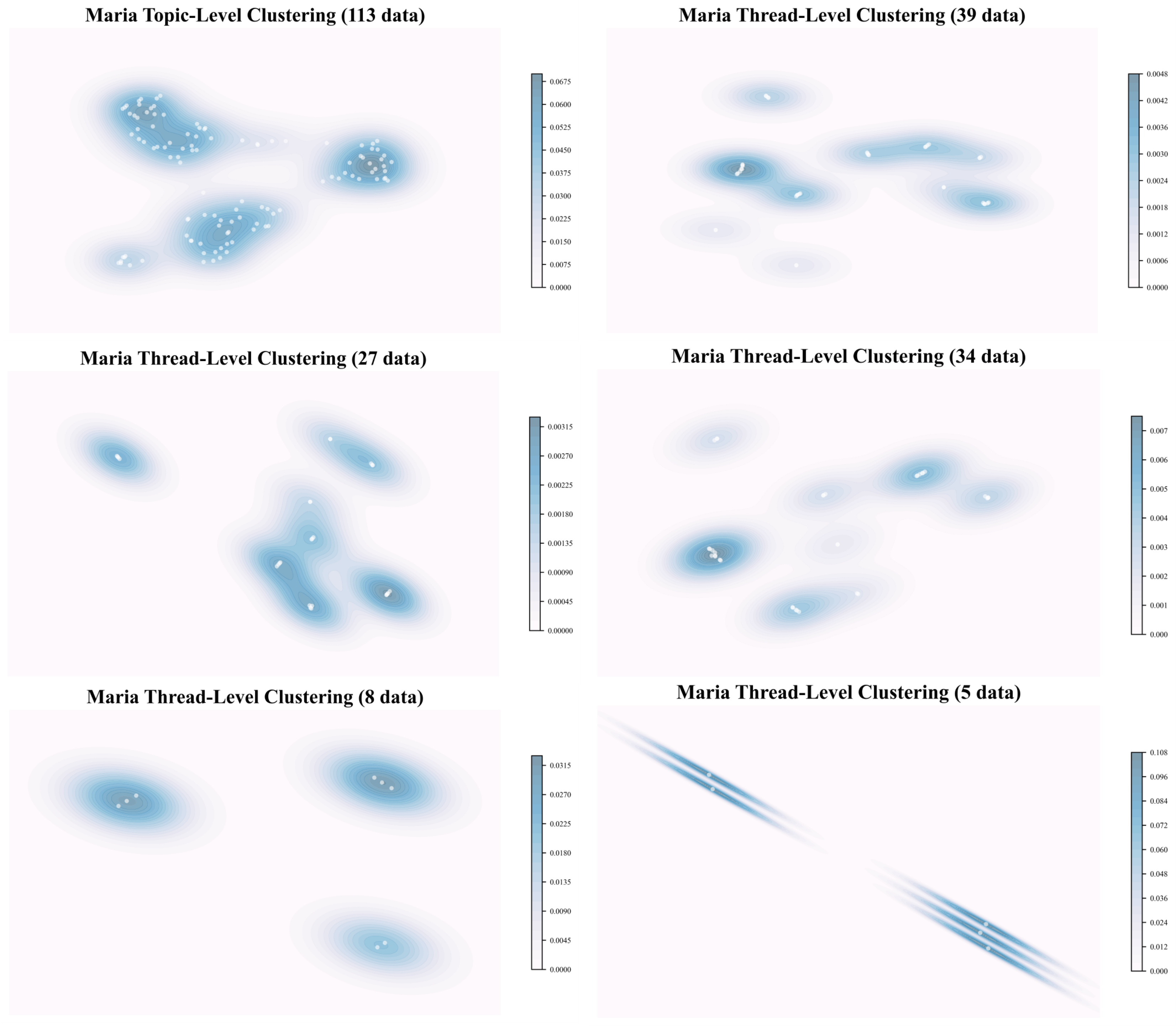} 
        \caption{User B (Maria) Clustering Results}
        \label{fig:cluster22}
    \end{subfigure}
    \caption{An example of clustering result of GPT-4.1-mini backbone.}
    \label{fig:cluster2}
\end{figure}

Moreover, Fig.~\ref{fig:cluster1} and Fig.~\ref{fig:cluster2} visualize the clustering in systems memory consolidation process for a randomly selected dialogue between John and Maria, utilizing GPT-4o-mini and GPT-4.1-mini backbones, respectively. These diagrams represent the memory embedding space as a density distribution via Gaussian Kernel Density Estimation (KDE). Each visualization incorporates topic-level clusters alongside the top five thread-level segments to illustrate the system's organizational hierarchy.

\subsection{Ablation Study}
To evaluate the effectiveness of the individual components within TraceMem, we conduct a series of ablation studies. The ablation baselines are defined as follows:
\begin{itemize}[leftmargin=*, nosep]
    \item \textbf{Epidodic Memory Only} This configuration retains only the top $K=10$ episodic memories retrieved via vector similarity. It is designed to evaluate the performance of the system when relying solely on deductive topic segmentation and episodic summarization.
    \item \textbf{Episodic Memory + Semantic Representation (20)} This baseline augments the aforementioned episodic memories with 20 semantic representations. These representations are extracted during the short-term memory processing stage. This follows the ``episodic + semantic'' paradigm, a common practice adopted in many existing studies without memory consolidation.
    \item \textbf{Episodic Memory + Semantic Representation (40)} This baseline increases the number of semantic representations to 40. It is intended to test whether further increasing the density of semantic information continues to yield performance gains.
    \item \textbf{Without Agentic Search} This configuration deactivates the agentic search mechanism while preserving the complete memory formulation process of TraceMem, thereby enabling a direct assessment of the mechanism's contribution to the system's performance.
\end{itemize}

As shown in Table~\ref{tab:ablation_results}, the model utilizing only episodic memory achieved an overall score of 0.8429. Specifically, it scored 0.8954 on the SingleHop task, 0.8262 on the MultiHop task, 0.7664 on the Temporal task, and 0.6875 on the OpenDomain task. This pattern demonstrates a clear and progressive decline in performance as task complexity increases, establishing a baseline that highlights the limitations of a single memory module.

The introduction of semantic representations led to a systematic improvement in performance. With a 20 representations, the overall score rose to 0.8701. Expanding the representation to 40 dimensions further increased the overall score to 0.8792. These results confirm that enhancing semantic understanding helps in certain tasks, but also indicates a performance plateau in others, particularly in open-domain and temporal reasoning.
Removing the agentic search mechanism resulted in an overall score of 0.8844. This configuration achieved a notably high OpenDomain performance of 0.7188 and a Temporal score of 0.8349. However, this came at the cost of reduced capability in core reasoning tasks: the SingleHop score dropped to 0.9287 and the MultiHop score fell significantly to 0.8652. This illustrates that a simplified retrieval strategy impairs the model’s ability to perform complex, multi-step reasoning.


\section{Open Discussion}
Delving into the mechanics of human memory, some studies support the fact that memory is rewritten each time it is retreived~\cite{schwabe2014reconsolidation}. Essentially, it means that every act of recollection holds the potential to update information or trigger subtle reorganizations. This malleability is particularly intriguing for interactive agents, as an agent retrieves past experiences to inform current responses, its memory is not merely accessed but potentially reshaped, contributing to a dynamic where memory may evolve through active conversation and internal monologue, potentially leading to alterations over time.

Such rewriting could enable the system to learn what to prioritize and what to disregard. Much like the human brain, where seldom-accessed knowledge tends to decay, information that remains unvisited in long-term interactions may reflect aspects that no longer require reinforcement. Consequently, determining what to forget and at what rate emerges as a pivotal challenge. Therefore, memory rewriting upon retrieval and regulated forgetting may represent two crucial mechanisms for future development.

Looking ahead, we may develop agents endowed with diverse personality traits, ranging from irritable and easy-going to anxious or emotionally complex. What they prioritize or disregard could also stem from these intrinsic characteristics, thereby making the dynamics of memory even more intriguing.

But even given such memory, should an agent utilize it at all times? Determining when to invoke memory and when to abstain may be a critical focus for future research. Even in conversations with the same user, there are instances where a task can be performed perfectly well without accessing past data. In fact, incorporating excessive retrieved memories could potentially introduce interference, hindering the quality of the response. Accordingly, an ideal memory system should evolve into a dynamic cognitive partner that can judiciously activate or suppress memories during exchanges.

\section{Conclusion}
In this paper, we introduced a cognitive-inspired memory system TraceMem. Through a three-stage pipeline, the system transforms conversational traces into structured user memory cards and leverages an agentic aearch mechanism for precise source attribution. Evaluations on the LoCoMo benchmark show that TraceMem significantly outperforms state-of-the-art baselines, particularly in MultiHop and Temporal reasoning tasks.

\bibliographystyle{ACM-Reference-Format}
\bibliography{sample-base}

\clearpage
\appendix
\onecolumn
\section{Prompt Templates}
We provide the full suite of prompt templates used in TraceMem, including those for deductive episodic segmentation, episodic summarization, user experience trace distillation, user card selection, agentic search, thread summarization, response generation and the LLM-as-a-judge prompt provided by Nemori.

\begin{figure}[h]
\centerline{\includegraphics[width=\textwidth,height=0.85\textwidth]{ts.pdf}}
\label{ts_prompt}
\end{figure}

\begin{figure}[h]
\centerline{\includegraphics[width=0.98\textwidth,height=0.6\textwidth]{es.pdf}}
\label{pm_prompt}
\end{figure}

\begin{figure}[h]
\centerline{\includegraphics[width=0.98\textwidth,height=0.62\textwidth]{pm.pdf}}
\label{es_prompt}
\end{figure}

\begin{figure}[h]
\centerline{\includegraphics[width=0.94\textwidth,height=0.6\textwidth]{uc.pdf}}
\label{uc_prompt}
\end{figure}

\begin{figure}[h]
\centerline{\includegraphics[width=0.94\textwidth,height=0.6\textwidth]{ap.pdf}}
\label{uc_prompt}
\end{figure}

\begin{figure}[h]
\centerline{\includegraphics[width=0.9\textwidth,height=\textwidth]{sp.pdf}}
\label{uc_prompt}
\end{figure}

\begin{figure}[h]
\centerline{\includegraphics[width=0.88\textwidth,height=0.5\textwidth]{test.pdf}}
\label{uc_prompt}
\end{figure}

\begin{figure}[h]
\centerline{\includegraphics[width=0.9\textwidth,height=0.62\textwidth]{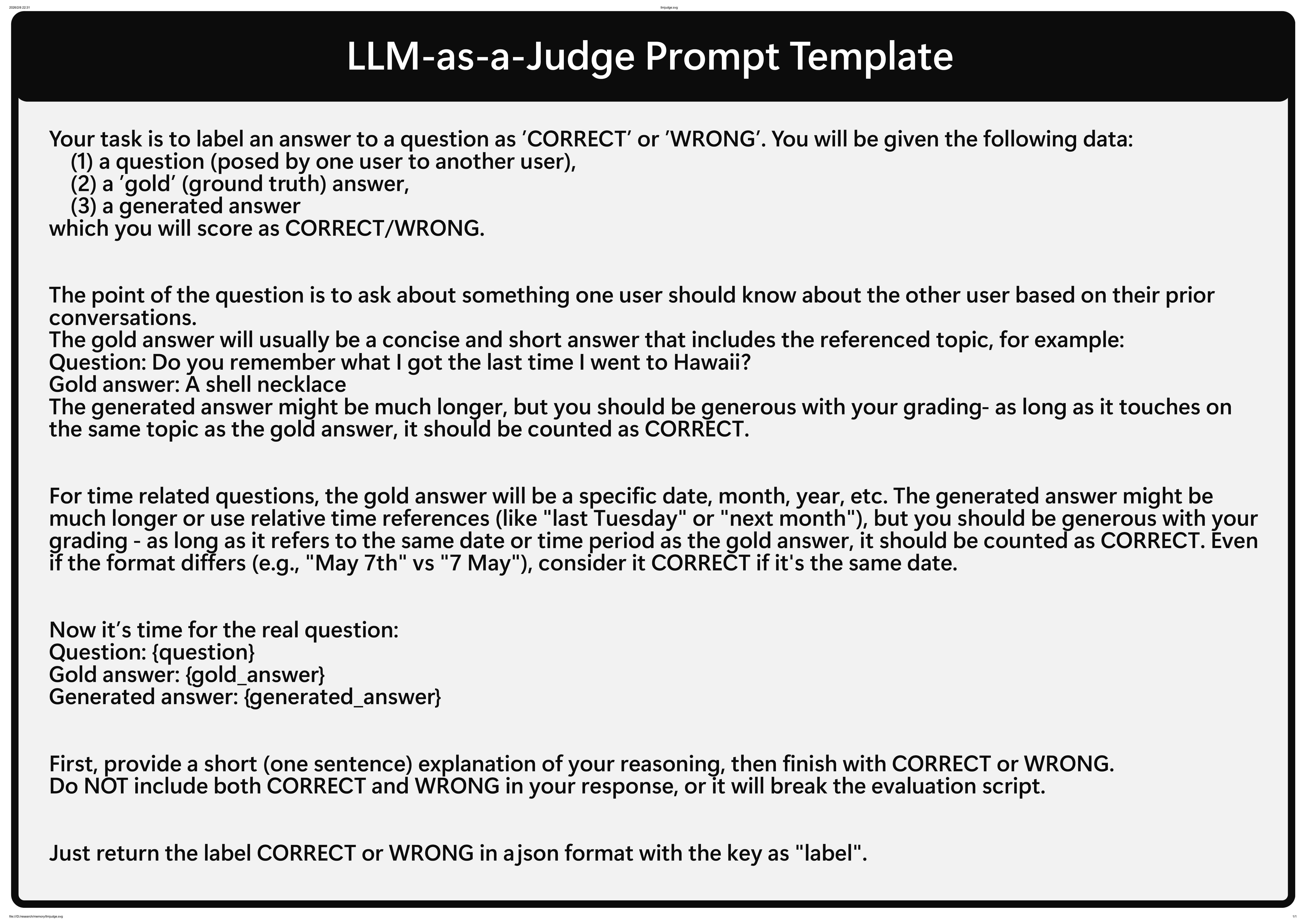}}
\label{uc_prompt}
\end{figure}

\end{document}